\documentclass[runningheads]{llncs}
\usepackage{graphicx}
\usepackage{subcaption}

\begin{document}
\title{To Ensemble or Not: Assessing Majority Voting Strategies for Phishing Detection with Large Language Models\thanks{Supported by the Maroun Semaan Faculty of Engineering (MSFEA) at the American University of Beirut (AUB)}}
\titlerunning{Assessing Majority Voting Strategies for Phishing Detection with LLMs}
%
\author{Fouad Trad \and
Ali Chehab }
\authorrunning{F. Trad and A. Chehab}
%
\institute{Electrical and Computer Engineering\\ American University of Beirut\\ Beirut, Lebanon\\
\email{fat10@mail.aub.edu}, \email{chehab@aub.edu.lb}} 
\maketitle              
\begin{abstract}
The effectiveness of Large Language Models (LLMs) significantly relies on the quality of the prompts they receive. However, even when processing identical prompts, LLMs can yield varying outcomes due to differences in their training processes. To leverage the collective intelligence of multiple LLMs and enhance their performance, this study investigates three majority voting strategies for text classification, focusing on phishing URL detection. The strategies are: 1) a prompt-based ensemble, which utilizes majority voting across the responses generated by a single LLM to various prompts 2) a model-based ensemble, which entails aggregating responses from multiple LLMs to a single prompt; and 3) a hybrid ensemble, which combines the two methods by sending different prompts to multiple LLMs and then aggregating their responses. Our analysis shows that ensemble strategies are most suited in the cases where individual components—whether prompts or LLMs—exhibit equivalent performance levels. However, when there is a significant discrepancy in individual performance, the effectiveness of the ensemble method may not exceed that of the highest-performing single LLM or prompt. In such instances, opting for ensemble techniques is not recommended.

\end{abstract}

\section{Introduction}
Large Language Models (LLMs) have become instrumental in advancing tasks within natural language processing (NLP), offering unparalleled capabilities in text comprehension and generation \cite{chang2023survey,wei2022emergent}. However, the performance of LLMs is significantly reliant on the specificity and design of the input prompts \cite{marvin2023prompt}. This dependence frequently leads to inconsistent results across different LLMs when processing the same prompt, a consequence of their diverse training methodologies.

The current literature predominantly addresses enhancements to individual LLMs through architectural improvements \cite{raiaan2024review}, diversified training data \cite{moore2010intelligent}, prompt engineering \cite{jiang2022promptmaker}, and fine-tuning \cite{ding2023parameter}. However, there is a scarcity of research focused on ensemble methods that synergize the strengths of multiple LLMs to boost performance. This research gap presents an opportunity to examine how ensemble strategies might reduce the variance in LLM responses and increase their predictive accuracy.

This study aims to fill this gap by introducing and evaluating three majority voting ensemble strategies tailored for text classification, with a special focus on phishing URL detection. The strategies explored are: (1) a prompt-based ensemble that uses majority voting across the responses elicited by a single LLM from various prompts; (2) a model-based ensemble that combines responses from multiple LLMs given a single prompt;  and (3) a hybrid approach that merges the two preceding strategies by sending multiple prompts to multiple LLMs and aggregating the results.

Notably, we experiment with three types of prompts—zero-shot, one-shot, and two-shot—across five LLMs: GPT-3.5-Turbo, GPT-4, Gemini-1.0-pro, PaLM 2, and LLaMA 2 (70B), implementing the aforementioned ensemble techniques. Our analysis reveals that ensemble strategies yield benefits when the performance of the individual components within the ensemble is comparable. However, in instances where there is a significant disparity in performance among these components, the output of the ensemble is unlikely to surpass that of the single highest-performing component, rendering the ensemble technique ineffective.

In summary, the principal contributions of this paper are as follows:

\begin{itemize}
    \item The introduction of three innovative majority voting ensemble strategies to enhance text classification with LLMs.
    \item Empirical validation of these strategies within the context of phishing URL detection, highlighting their strengths and limitations.
    \item The provision of practical insights for the application of ensemble methods in LLMs, aiming to optimize performance.
\end{itemize}

The remainder of this paper is organized as follows: Section 2 discusses the background of LLMs and the particular challenges our research seeks to address. Section 3 reviews the relevant literature. In Section 4, the proposed ensemble strategies are introduced. Section 5 describes the experimental setup and the experiments used to evaluate these strategies. Section 6 analyzes the results, discussing the implications and providing guidance on the applicability of the ensemble techniques. Section 7 concludes the work and proposes potential avenues for future research in this domain.

\section{Background and Preliminaries}
This section provides essential background information on key topics that form the foundation of this study.
\subsection{Large Language Models (LLMs)}
LLMs are neural-network-based models that excel at processing human language, benefiting significantly from their extensive scale and depth. The introduction of the transformer architecture by Vaswani et al. \cite{vaswani2017attention}, with its innovative self-attention mechanisms, has been a cornerstone in the development of LLMs. This architecture has led to significant improvements in various tasks, including translation and sentiment analysis. A major leap in the capabilities of LLMs has been demonstrated through the advancements in transformer-based models, notably exemplified by OpenAI's GPT series \cite{brown2020language}.

Among these, ChatGPT, a variant of GPT-3 optimized for conversational tasks, initiated a crucial evolution toward LLMs designed for a wide array of applications \cite{ray2023chatgpt}. Following the success and the trends set by ChatGPT, other significant LLMs have emerged, each contributing unique strengths to the field. LLaMA \cite{touvron2023llama}, introduced by Meta, PaLM \cite{anil2023palm} and Gemini \cite{team2023gemini}, by Google, are examples of such models that have adopted and extended the transformer architecture, further pushing the boundaries of what LLMs can achieve. These models, similar to ChatGPT, showcase the trend towards employing these sophisticated neural networks as off-the-shelf solutions for diverse systems and applications, thereby eliminating the need for extensive model training and maintenance from the end users' perspective.

\subsection{Phishing URL Detection}
Phishing attacks continue to pose challenging threats to global cybersecurity efforts. Attackers frequently employ deceptive URLs that mimic legitimate websites, utilizing misleading domain names, incorporating trusted brands, and using visually similar characters, or homoglyphs, to dupe users \cite{sern2020phishgan}. They further complicate detection by exploiting valid TLS certificates and brand logos, as well as by using URL shortening and redirection techniques \cite{james2013detection}. Given the sophistication and constant evolution of these tactics, there is a pressing need for advanced URL analysis methods. This study investigates the use of ensemble LLMs for text classification as a means to enhance phishing URL detection.

\subsection{Ensemble Strategies}
Ensemble Strategies have emerged as a powerful approach to improving the performance and reliability of predictive models. By combining the outputs of multiple models, ensemble methods can leverage diverse perspectives and knowledge bases to yield more accurate predictions \cite{dietterich2000ensemble}. Majority Voting is a simple yet effective ensemble technique where the final output is determined by the majority decision of multiple models. This approach assumes that each model in the ensemble has an independent chance of being correct, and by aggregating their decisions, the ensemble can achieve higher accuracy than any single model alone \cite{kuncheva2003limits}. This study aims to introduce and evaluate three majority voting approaches for LLMs in text classification tasks, with a focus on detecting phishing URLs.

\section{Related Work}
The utilization of LLMs has showcased their revolutionary abilities in a variety of NLP tasks \cite{kaddour2023challenges}. Among these tasks, text classification has emerged as a critical area where LLMs, including foundational models like BERT and GPT, demonstrate substantial potential. These models have been instrumental in advancing our understanding of nuanced language patterns, significantly enhancing analytical capabilities in classifying texts \cite{gonzalez2020comparing,pawar2022comparison}. This evolving landscape has further expanded with the introduction of chat-based LLMs which, through innovative prompt engineering and response parsing techniques, are being tailored for more specialized tasks \cite{zhao2023chatagri}. Building on this momentum, this study focuses on using chat-based LLMs for phishing URL detection, showcasing how these advanced models can be leveraged to address specific and pressing cybersecurity challenges.

Research in phishing URL detection has progressed from traditional blacklist methods \cite{sheng2009empirical} to more advanced machine learning-based approaches \cite{abu2007comparison}. In machine-learning-based phishing detection, significant emphasis is placed on URL scrutiny. These methods analyze URL features such as length, special character presence, and subdomain usage \cite{sahingoz_machine_2019,wei_accurate_2020,zouina_novel_2017}. Lexical and token analysis are crucial for deconstructing URL structures to identify phishing attempts. Machine learning algorithms like SVM, decision tree, and Random Forest play a pivotal role in categorizing URLs based on these attributes \cite{mahajan_phishing_2018,ahammad_phishing_2022}. The integration of deep learning techniques enhanced phishing URL detection by introducing character-level deep neural networks for this purpose \cite{le_urlnet_2018,jiang2018deep}. Building on these advancements, recent studies have introduced the use of prompt-engineered LLMs, achieving promising results. However, these approaches did not surpass the performance of fine-tuned models \cite{trad2024evaluating,trad2024prompt}. Notably, the studies were conducted within the constraints of a single-model, single-prompt framework, indicating that exploring ensemble methods could offer significant improvements in accuracy and reliability.

Ensemble methods like majority voting, bagging, and boosting improve performance by aggregating multiple models' predictions \cite{dietterich2000ensemble}. Although these methods have seen widespread use with simpler models, their application in NLP and text classification, especially with recent LLM advancements, has been less investigated.

The literature indicates a research gap in applying ensemble methods to harness multiple LLMs' collective strengths. While the individual text classification abilities of LLMs are well-established, there is limited research on their effective combination to boost performance. This gap underlines the necessity for dedicated studies on developing and assessing ensemble strategies for improved reliability and accuracy.

This study seeks to bridge this gap by introducing and evaluating three majority voting ensemble strategies, specifically designed for text classification tasks within the context of phishing URL detection.

\section{Methodology}
\begin{figure}[h]
    \centering
    
    
    \begin{subfigure}[b]{0.48\textwidth}
        \centering
        \includegraphics[width=\textwidth]{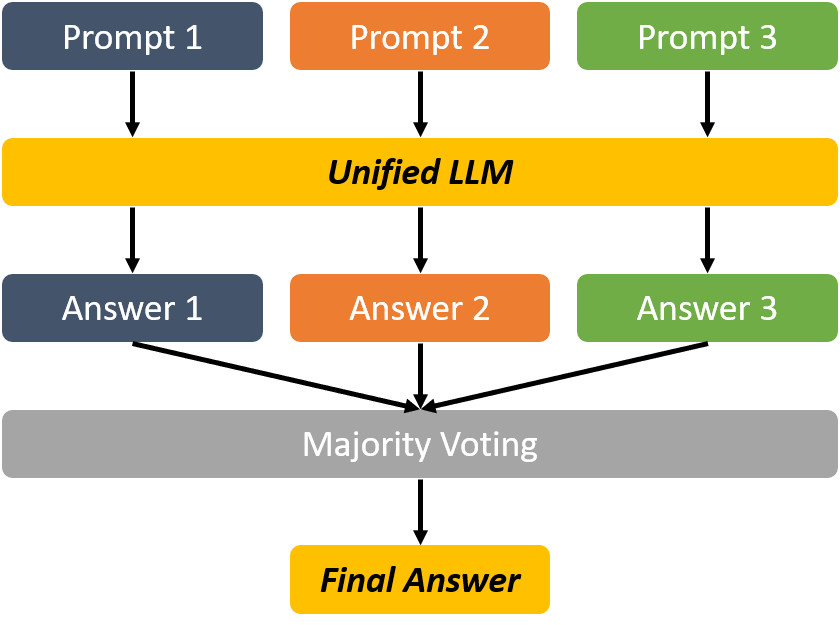}
        \caption{Prompt-based Ensemble}
        \label{fig:prompt_ensemble}
    \end{subfigure}
    \hfill 
    \begin{subfigure}[b]{0.48\textwidth}
        \centering
        \includegraphics[width=\textwidth]{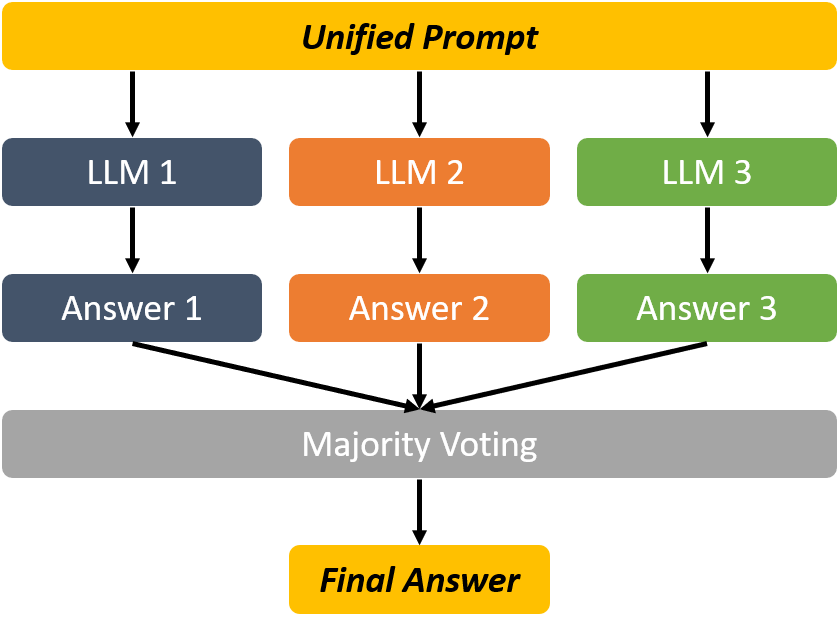}
        \caption{Model-based Ensemble}
        \label{fig:model_ensemble}
    \end{subfigure}%
    
    \vspace{10mm} 
    \begin{subfigure}[b]{\textwidth}
        \centering
        \includegraphics[width=\textwidth]{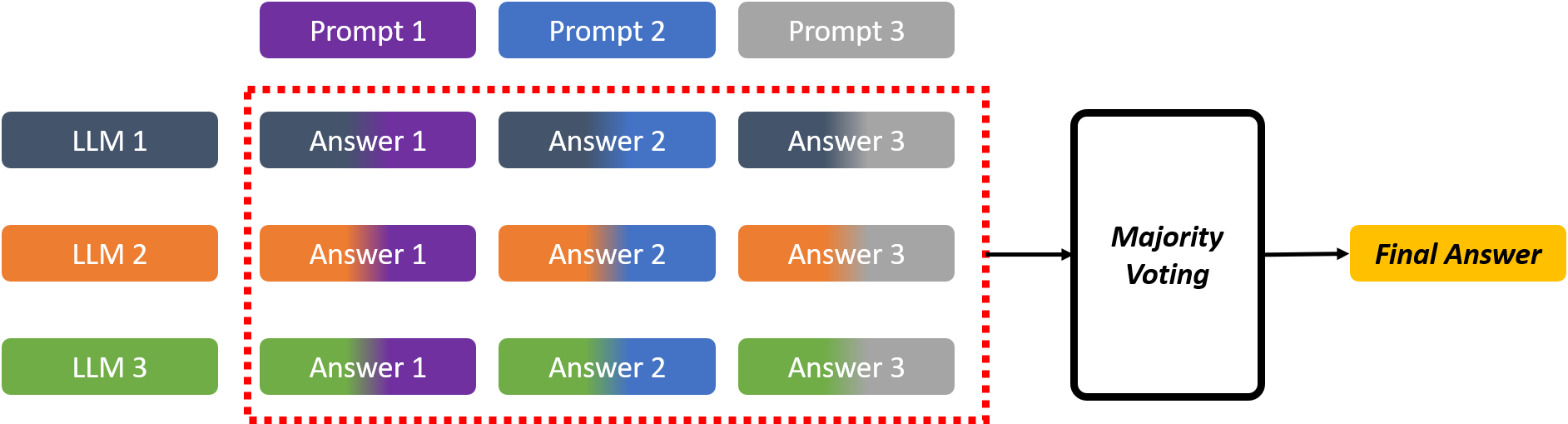}
        \caption{Hybrid Ensemble}
        \label{fig:hybrid_ensemble}
    \end{subfigure}

    \caption{Proposed ensemble methods}
    \label{fig:ensembles}
\end{figure}

This section outlines the three proposed ensemble strategies specifically designed to improve phishing URL detection with LLMs. These strategies, detailed in Figure \ref{fig:ensembles}, are as follows:

\begin{itemize}
    \item \textbf{Prompt-based Ensemble:} This method involves presenting a single LLM with multiple variations of prompts, each requesting the classification of the same URL. The responses are then aggregated through majority voting to determine the final classification. Figure \ref{fig:prompt_ensemble} illustrates this approach with an LLM receiving three different prompts.
    \item \textbf{Model-based Ensemble:} Here, responses from multiple LLMs to the same prompt are combined. Each model independently evaluates the prompt, and their decisions are aggregated via majority voting to arrive at the final prediction. Figure \ref{fig:model_ensemble} depicts this strategy with three LLMs analyzing a single prompt.
    \item \textbf{Hybrid Ensemble:} A synthesis of the prior strategies, the hybrid method sends various prompts to multiple LLMs and compiles their responses using majority voting. This approach aims to exploit simultaneously the benefits of model diversity and prompt variation. Figure \ref{fig:hybrid_ensemble} demonstrates this process for three LLMs, each receiving three distinct prompts.
\end{itemize}

\section{Experiments}
\subsection{Experimental Setup}
\subsubsection{Dataset:} 
The primary dataset for this project is derived from the PhishStorm dataset \cite{marchal2014phishstorm}, comprising 96,018 URLs, equally divided into 48,009 legitimate and 48,009 phishing URLs. To facilitate a focused and manageable evaluation of the proposed ensemble strategies across different LLMs and prompts, we selected a subset of 1,000 samples from the PhishStorm dataset for testing. This subset maintains the dataset's balanced structure, with an equal distribution of 500 phishing and 500 legitimate URLs.

\subsubsection{LLMs:}
To assess the effectiveness of ensemble strategies in phishing URL detection, we selected a diverse array of advanced, chat-based LLMs capable of being queried with prompts. This assortment includes a range of models showcasing the current breadth of capabilities in language understanding and processing. The models utilized in this study are: GPT-3.5-Turbo \cite{ye2023comprehensive}, GPT-4 \cite{achiam2023gpt}, Gemini 1.0 Pro \cite{team2023gemini}, LLaMA 2 (70B) \cite{touvron2023llama}, and PaLM 2 \cite{anil2023palm}.

\subsubsection{Prompts:}
\begin{figure}[h]
\centering
\includegraphics[width=\linewidth]{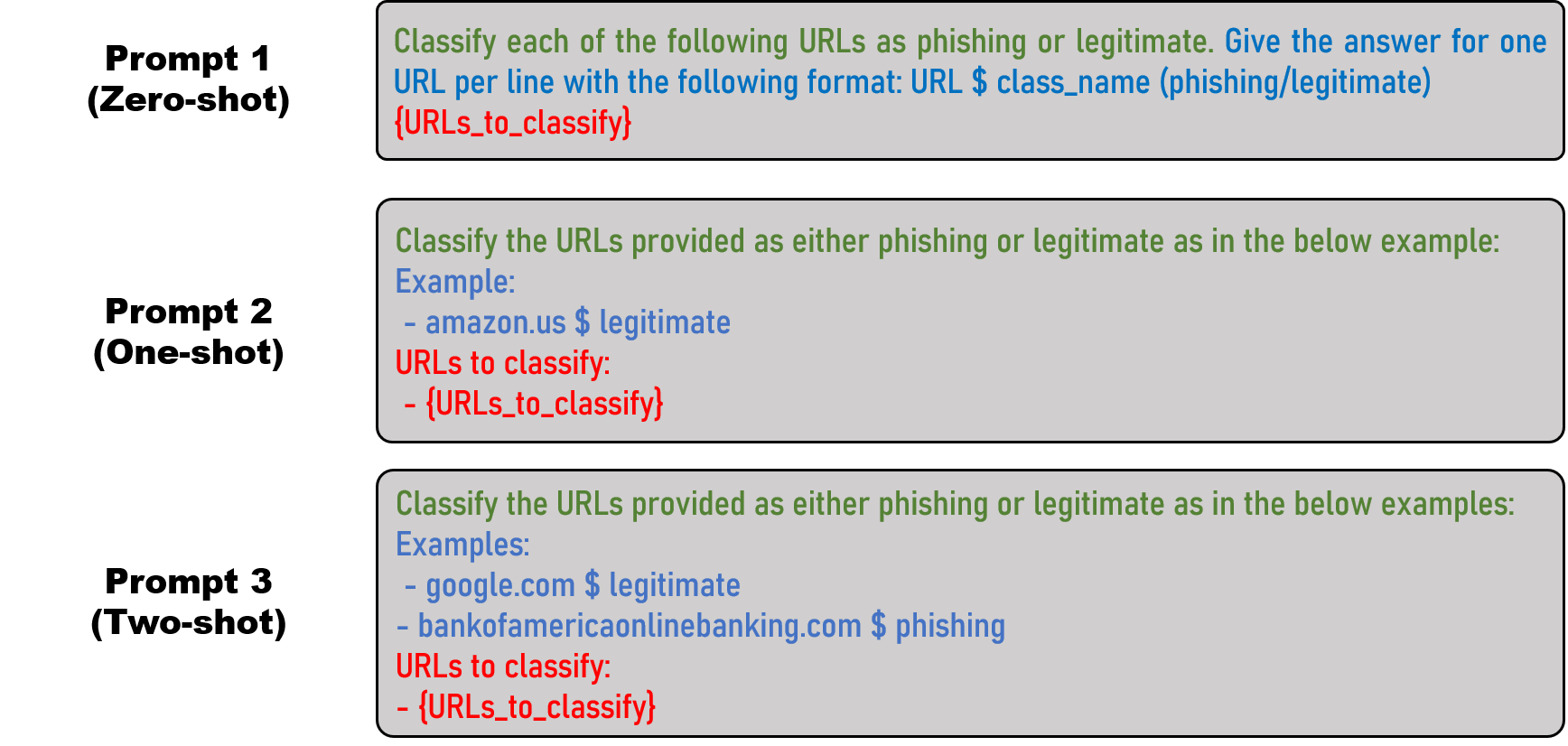}
\caption{Zero-shot, one-shot, and two-shot prompts used to classify URLs as phishing or legitimate}
\label{fig:prompts}
\end{figure}

To assess the performance of the selected LLMs in phishing URL detection, specific prompts were crafted to guide the models to classify URLs as either phishing or legitimate. The dataset was divided into batches of 50 samples for testing, with each LLM tasked with classifying all URLs in a batch based on a single prompt. The prompts utilized are as follows:
\begin{itemize}
    \item \textbf{Zero-shot:} This prompt directs the LLM to classify each URL in a list without any previous examples and specifies the response format for easy classification.
    \item \textbf{One-shot:} This prompt provides one example to demonstrate the classification task and output format before the model is asked to classify new URLs.
    \item \textbf{Two-shot:} This prompt gives two examples — one phishing and one legitimate — to set a precedence before the model classifies new URLs.
\end{itemize}
Details of these prompt structures, including their precise wording and format, are illustrated in Figure \ref{fig:prompts}.

\subsubsection{Evaluation Framework:}
The performance of each ensemble strategy was assessed using classification metrics, such as accuracy and F1-score. These metrics offered insights into how effectively the ensemble strategies could classify URLs as phishing or non-phishing. We compared these results with the baseline performance of the individual components of the ensemble, enabling a clear evaluation of the benefits introduced by the majority voting methods.

\subsection{Individual LLMs Performance}
\begin{figure}[h]
\centering
\includegraphics[width=\linewidth]{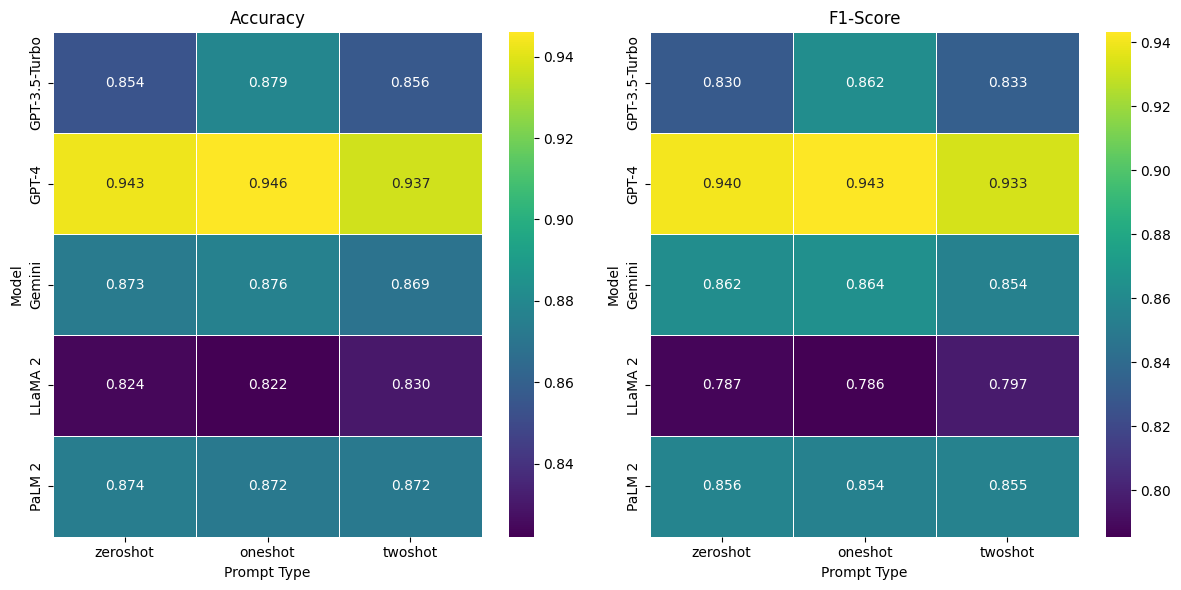}
\caption{Individual LLM performance for each prompt}
\label{fig:individualmodels}
\end{figure}

 As a start, we evaluate each LLM's performance using zero-shot, one-shot, and two-shot prompts. The individual performance of LLMs is necessary to establish baseline capabilities that ensemble strategies are expected to bypass. The heat-map depicted in Figure \ref{fig:individualmodels} visually conveys the accuracy and F1-score for each model-prompt pairing.

GPT-4 outperforms other models, achieving the highest accuracy of 0.946 with the one-shot prompt and an F1-score of 0.943. In contrast, LLaMA 2 is at the lower performance spectrum, attaining its highest accuracy of 0.830 and an F1-score of 0.797 when given the two-shot prompt. The performance of Gemini-1.0-Pro, PaLM 2, and GPT-3.5-Turbo occupies the middle ground. 

Gemini demonstrates negligible variation in performance across prompts, suggesting robustness to prompt complexity with consistent accuracy and F1-scores. PaLM 2 shows similar consistency for both metrics. Conversely, GPT-3.5-Turbo presents a variable performance, with a notable accuracy improvement from 0.854 for zero-shot to 0.879 for one-shot, and a subsequent decline to 0.856 in the two-shot prompt. F1-scores also reflect this pattern, peaking at 0.862 for one-shot before dropping to 0.833 for two-shot prompts. 

Understanding the individual model performances is crucial for determining the effectiveness of ensemble strategies, particularly when they can surpass the individual performances of the models involved.

\subsection{Prompt-based Ensemble}
\begin{figure}[h]
\centering
\includegraphics[width=\linewidth]{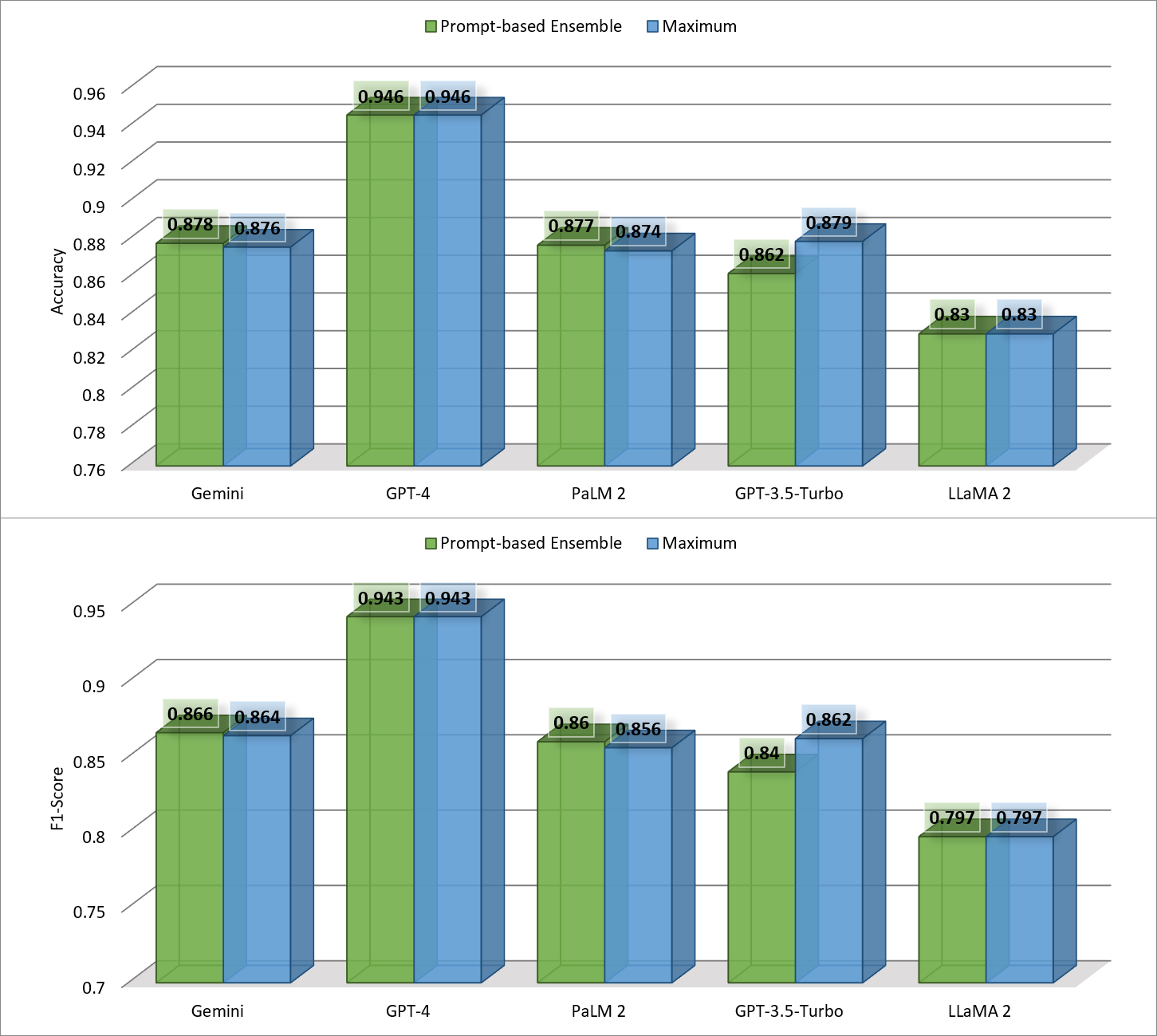}
\caption{Comparison of prompt-based ensembling with the highest performing prompt for each LLM}
\label{fig:prompt_performance}
\end{figure}

Applying the prompt-based ensemble technique outlined in the Methodology section, we report on the performance of each LLM when the results from individual prompts are combined, comparing these to the highest performance achieved by any single prompt. The outcomes are illustrated in Figure \ref{fig:prompt_performance}. For all models, the prompt-based ensemble either improved or matched the best individual performance without ensemble. The only exception is GPT-3.5-Turbo, which experienced a minor decrease in performance due to the initial variable performance across prompts: zero-shot and two-shot prompts resulted in approximately 0.85 accuracy and an F1-score around 0.83, whereas the one-shot prompt yielded higher outcomes with 0.88 accuracy and an 0.86 F1-score. Since two out of the three prompts had lower performance, the ensemble tended to skew towards these lesser-performing prompts rather than achieving the optimal individual performance. This result demonstrates that prompt-based ensemble is more effective when the prompts exhibit similar performance levels for a given LLM.

\subsection{Model-based Ensemble}
The model-based ensemble approach was implemented to assess the impact of aggregating the predictive outputs of various LLMs. In this method, a single prompt is provided to all models, and their individual predictions are consolidated through majority voting for the final URL classification.

\begin{figure}[h]
\centering
\includegraphics[width=\textwidth]{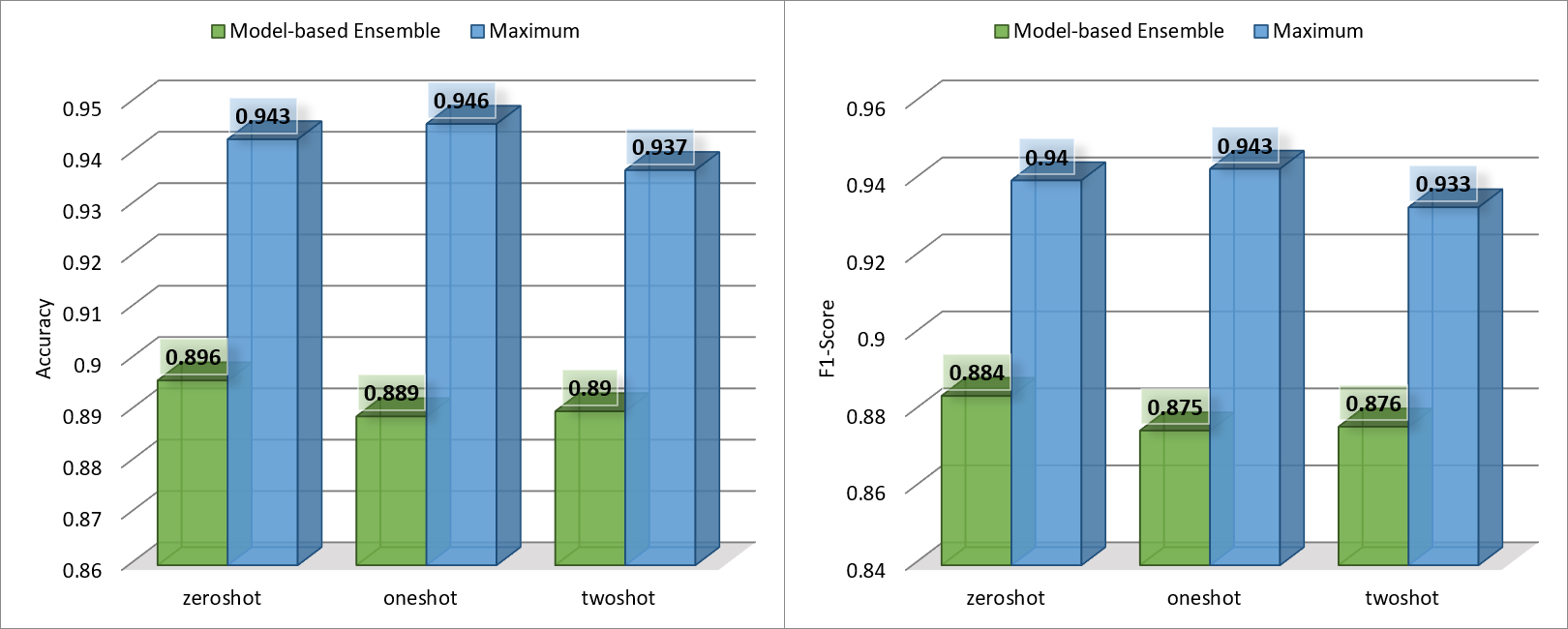}
\caption{Comparison of model-based ensembling across the five models, with the highest performing model for each prompt type}
\label{fig:model_performance}
\end{figure}

The outcomes of this experiment, depicted in Figure \ref{fig:model_performance}, illustrate the model-based ensemble's efficacy with the highest performance achieved by individual models for a given prompt type. Notably, the model-based ensemble failed to surpass the performance of the highest-performing model (GPT-4). This is attributed to GPT-4's significantly superior performance within the ensemble of five LLMs, resulting in a tendency for the ensemble's overall performance to lean towards the lower-performing models.

\begin{figure}[h]
\centering
\includegraphics[width=\textwidth]{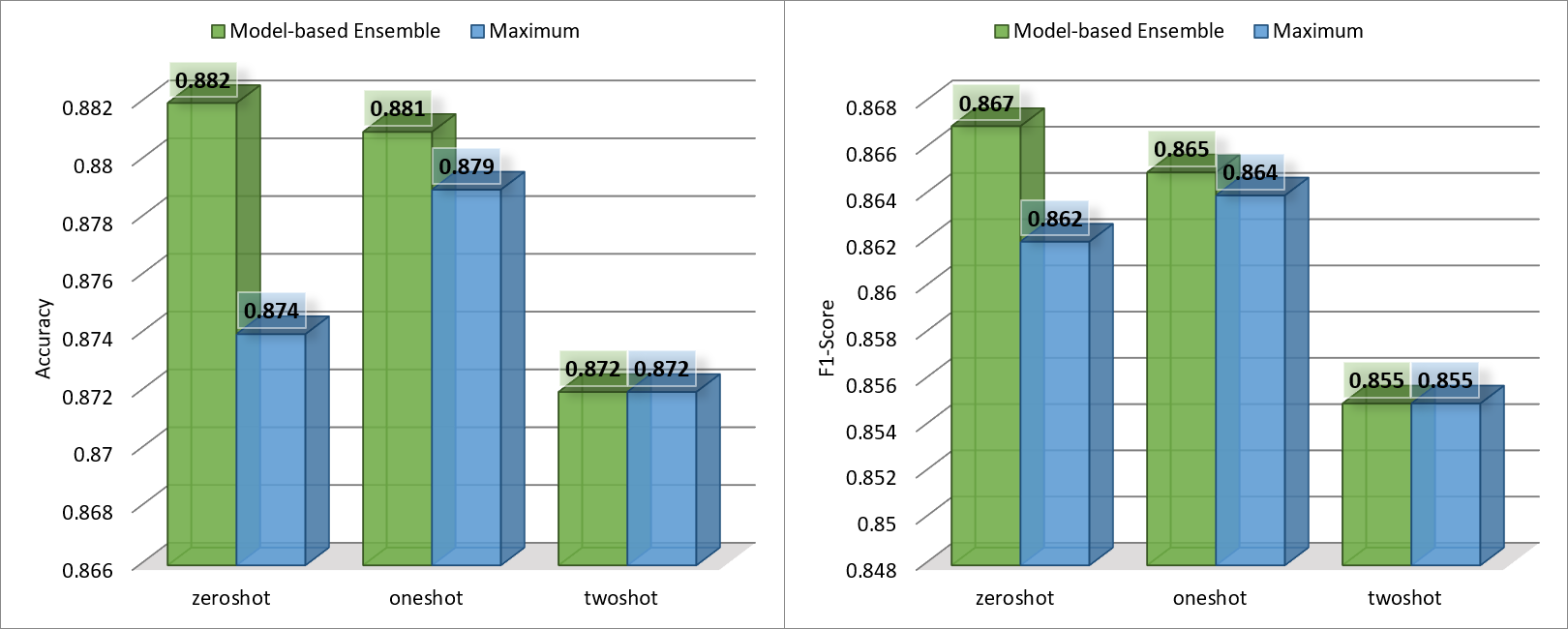}
\caption{Comparison of model-based ensembling among the similarly performing models, with the highest performing model for each prompt type}
\label{fig:improved_model_performance}
\end{figure}

In an effort to further investigate, a follow-up experiment was conducted, excluding both the highest (GPT-4) and lowest (LLaMA) performing models, focusing instead on ensemble models (Gemini, GPT-3.5-Turbo, and PaLM) with more uniform performance. The results, shown in Figure \ref{fig:improved_model_performance}, confirm that the ensemble models with similar performance indeed yields improved results across all prompt types compared to individual model outcomes. This underscores that model-based ensembling is most effective when the individual models exhibit comparable levels of performance. Otherwise, picking the highest-performing model will give better results.

\subsection{Hybrid Ensemble}
\begin{figure}[h]
\centering
\includegraphics[width=\textwidth]{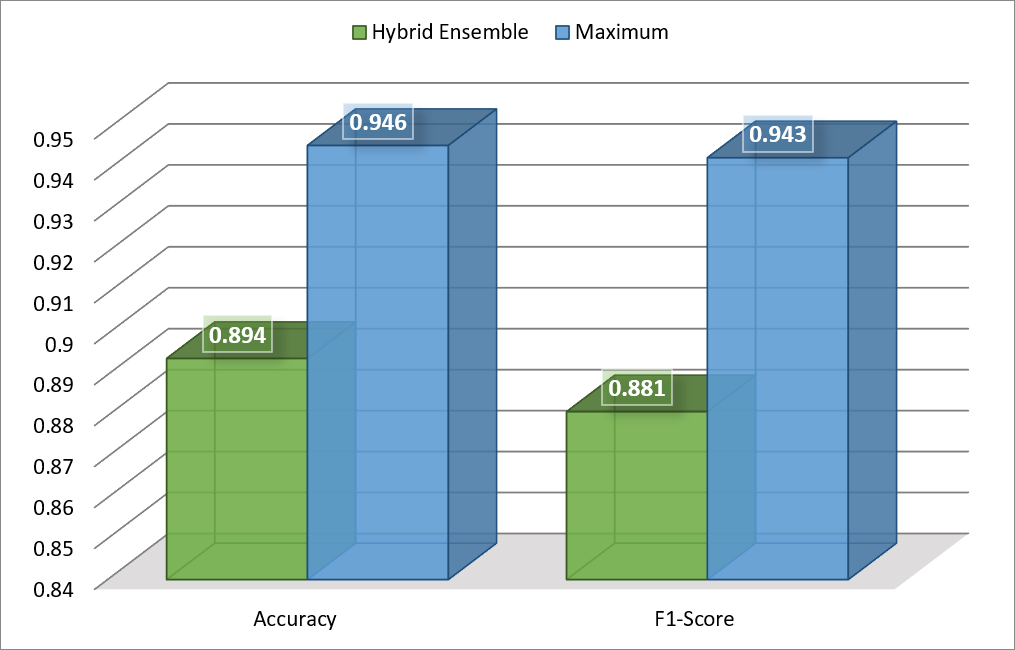}
\caption{Performance of the hybrid ensemble compared to the highest-performing individual model}
\label{fig:hybrid_performance}
\end{figure}

The hybrid ensemble methodology merges the principles of both prompt-based and model-based ensembling, utilizing various prompts across different LLMs. The predictive outcomes are then aggregated through majority voting. Including all five models in the ensemble revealed no improvement over the best individual model, consistently GPT-4, as demonstrated in Figure \ref{fig:hybrid_performance}. This lack of enhancement is due to GPT-4's significantly superior performance across all prompts within the ensemble, which skews the ensemble's overall performance towards the lower-performing models.

\begin{figure}[h]
\centering
\includegraphics[width=\textwidth]{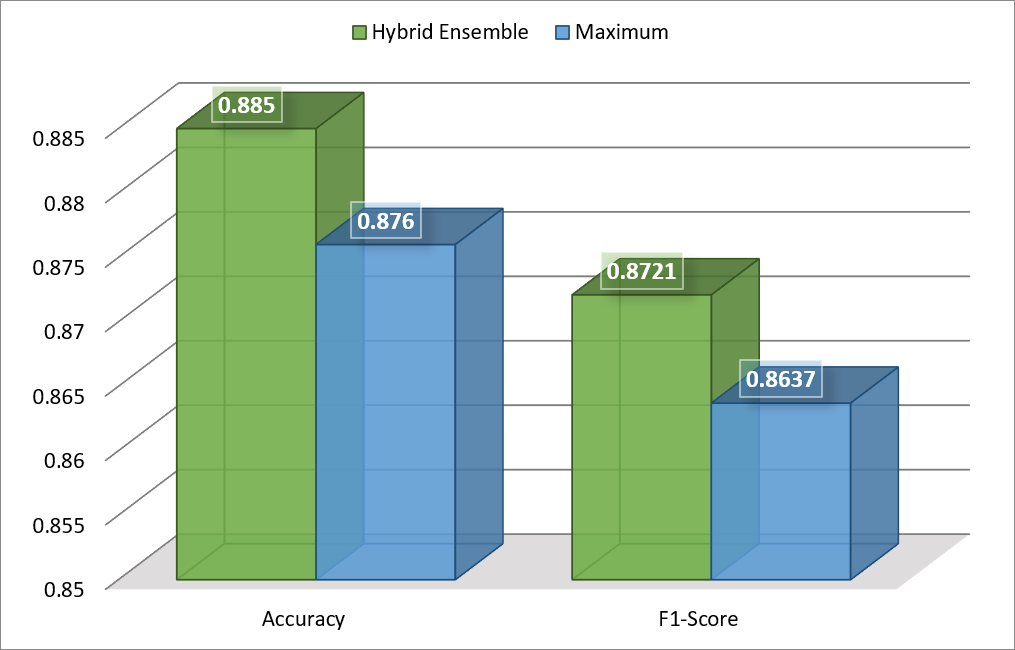}
\caption{Performance of hybrid ensemble among similarly performing models and prompts}
\label{fig:improved_hybrid_performance}
\end{figure}

In response, we excluded models with the most disparate performance: the highest-performing GPT-4 and the lowest-performing LLaMA, as well as models displaying significant performance variability across prompts, such as GPT-3.5-Turbo. This adjustment led to a more balanced ensemble comprising only Gemini and PaLM. This refined ensemble displayed a significant improvement, as illustrated in Figure \ref{fig:improved_hybrid_performance}, confirming the efficacy of hybrid ensembling when integrating models and prompts with similar performance levels. 

\section{Discussion}
The decision to employ ensembling strategies with LLMs must be informed by a clear understanding of when such approaches are likely to enhance performance.  Our experiments have shed light on the scenarios in which ensembling is advantageous. Importantly, ensembling is most effective when the participating models perform at similar levels.

When models exhibit widely varying levels of performance,  as seen with the inclusion of GPT-4 with other LLMs, the ensemble's outcome may be disproportionately affected by the higher-performing model, leading to no real benefit from the combination. In contrast, when models with comparable performance are ensembled, as demonstrated by the refined approach excluding GPT-4 and LLaMA, a notable improvement is witnessed. This suggests that the benefits of ensembling are more pronounced when the collective decision-making process is not dominated by a single model's output.

Considering the limited pool of LLMs available for ensembling and the notable variability in their performance, careful consideration is essential in applying ensembling strategies. We recommend using a portion of the data to assess the performance of both individual models and their ensembles before deploying ensemble LLMs in a production environment. This preliminary evaluation will help determine the most effective ensemble configuration and ensure that the integration of multiple LLMs augments the system's overall performance, rather than inadvertently diminishing it.

\section{Conclusion}
In conclusion, this study explored the viability of using ensemble methods with LLMs for the task of phishing URL detection. The findings have emphasized that while ensembles can be powerful, their success is contingent on the performance parity of the individual models. We demonstrated through experiments that ensembling yields improvements when models have relatively similar performance but may not be advantageous when a single model's performance is significantly better or lower than others. This research contributes to the broader understanding of how to strategically combine LLMs to leverage their collective intelligence effectively. 

There are several promising avenues for future research. One immediate area is the exploration of dynamic ensembling techniques that can adaptively select models based on the nature of the task or the specific input data. This could potentially overcome the limitations of fixed ensembles when faced with diverse types of data or tasks.

Further investigations could also focus on developing more sophisticated voting mechanisms beyond the simple majority rule, potentially incorporating weighted voting systems that consider the confidence levels or past performance of individual models.

Finally, there is a need for larger-scale studies that incorporate a greater variety of LLMs, including those trained on specialized datasets, to examine the effects of ensembling across a broader spectrum of language understanding capabilities.

%
%
%
\bibliographystyle{splncs04}
\bibliography{mybibliography}
%




\end{document}